\documentclass[conference]{IEEEtran}
\IEEEoverridecommandlockouts
\usepackage{cite}
\usepackage{amsmath,amssymb,amsfonts}
\usepackage{algorithmic}
\usepackage{graphicx}
\usepackage{textcomp}
\usepackage[table]{xcolor}
\usepackage{booktabs} 
\usepackage{lipsum}
\usepackage{soul,color} 
\usepackage{gensymb} 
\usepackage{multirow}
\usepackage{url}
\usepackage{array}
\usepackage{comment}
\usepackage[T1]{fontenc}
\def\BibTeX{{\rm B\kern-.05em{\sc i\kern-.025em b}\kern-.08em
    T\kern-.1667em\lower.7ex\hbox{E}\kern-.125emX}}

\usepackage[ruled]{algorithm2e} 

\newcolumntype{P}[1]{>{\centering\arraybackslash}p{#1}}
\SetAlFnt{\small}
\SetAlCapFnt{\small}
\SetAlCapNameFnt{\small}
\SetAlCapHSkip{0pt}

\definecolor{mitred}{rgb}{0.64, 0.12, 0.20}
\definecolor{forestgreen}{rgb}{0.0, 0.27, 0.13}
\definecolor{auburn}{rgb}{0.43, 0.21, 0.1}

\newcommand{\yi}[1]{\textcolor{red}{[Yi: #1]}}

\IEEEoverridecommandlockouts
\IEEEpubid{\makebox[\columnwidth]{ 979-8-3503-5067-8/24/\$31.00~\copyright2024 IEEE \hfill} 
\hspace{\columnsep}\makebox[\columnwidth]{ }}

\IEEEpubidadjcol

\begin{document}

\title{PlotMap: Automated Layout Design for Building Game Worlds}

\author{
\IEEEauthorblockN{Yi Wang, Jieliang Luo, Adam Gaier, Evan Atherton, Hilmar Koch}
\IEEEauthorblockA{Autodesk Research \\
\texttt{\{yi.wang, rodger.luo, adam.gaier, evan.atherton, hilmar.koch\}@autodesk.com}}
}

\maketitle

\begin{abstract}
World-building, the process of developing both the narrative and physical world of a game, plays a vital role in the game's experience. 
Critically-acclaimed independent and AAA video games are praised for strong world-building, with game maps that masterfully intertwine with and elevate the narrative, captivating players and leaving a lasting impression. 
However, designing game maps that support a desired narrative is challenging, as it requires satisfying complex constraints from various considerations.
Most existing map generation methods focus on considerations about gameplay mechanics or map topography, while the need to support the story is typically neglected. 
As a result, extensive manual adjustment is still required to design a game world that facilitates particular stories. 
In this work, we approach this problem by introducing an extra layer of {\em plot facility layout design} that is independent of the underlying map generation method in a world-building pipeline. 

Concretely, we define {\em (plot) facility layout tasks} as the tasks of assigning concrete locations on a game map to abstract locations mentioned in a given story (plot facilities), following spatial constraints derived from the story. We present two methods for solving these tasks automatically: an evolutionary computation based approach through Covariance Matrix Adaptation Evolution Strategy (CMA-ES), and a Reinforcement Learning (RL) based approach. We develop a method of generating datasets of 
facility layout tasks, create a gym-like environment for experimenting with and evaluating different methods, and further analyze the two methods with comprehensive experiments, aiming to provide insights for solving facility layout tasks. 



\end{abstract}

\begin{IEEEkeywords}
World Building,
Procedural Content Generation, Game Narrative, Reinforcement Learning, CMA-ES
\end{IEEEkeywords}

\maketitle


\section{Introduction}

Landscapes in digital games serve as more than just scenic backdrops; they interact intimately with the unfolding narrative, defining and shaping the player's experience. 
Arming designers with tools considering narrative in map design, allows them to create more cohesive and immersive games.
However, designing game maps is difficult, as it requires designers to consider varied qualities such as realistic topography \cite{smelik2009survey,kelly2017survey} and game playability \cite{Linden2014procedural} at the same time. Designing a map that supports a given story adds more constraints, making the problem even more challenging. 

While the need to support an underlying story is typically neglected in most existing map generation methods, some efforts have been made to develop the story first, then to codify it as plot points and graphs so that maps can be generated based on their relations~\cite{valls2013towards, hartsook2011toward}. However, as Dormans and Bakkes~\cite{dormans2011generating} pointed out, the principles that govern the design of the spatial and the narrative side of the game are different, and thus these two processes should be independent. Methods for generating game maps from stories can appear artificially contrived to fit a story, and it is not straightforward to combine these methods with those that also take into account game design and geographical considerations. As a result, designing a game world that facilitates a story requires extensive manual modification; and as the number of constraints scale, the challenge of designing a map that satisfies all of the constraints of the story can become intractable, if not impossible, for a designer to do by hand \cite{matsumoto2022elden}.

We approach this problem by introducing an extra layer of {\em plot facility layout design} that is independent to the underlying map generation method in a world-building pipeline. 

Our work is inspired by the philosophy behind the work from Dormans and Bakkes~\cite{dormans2011generating}, which distinguishes the abstract space defined by the story (referred to as missions) and the concrete space defined by the actual geometric layout of a game map. The story is accommodated by mapping the former into the latter. While  Dormans and Bakkes~\cite{dormans2011generating} focus on action adventure games with discrete ``scenes'' connected by ``passages'', we impose very little assumption on methods used for story and map generation, and in particular target workflows for modern open world games.

We introduce the concept of {\em plot facilities}, which are abstract locations mentioned in the given story. A set of constraints are derived from the story in terms of the spatial relationships between plot facilities and elements in the underlying map. Given an underlying map, we arrange the layout of the plot facilities on top of the map to satisfy the constraints. Our method is compatible with any map generation technique in the sense that the underlying map can be hand-crafted, procedurally generated, or even from a Geographic Information System (GIS) such as Google Maps.

\begin{figure*}
    \includegraphics[width=0.97\textwidth]{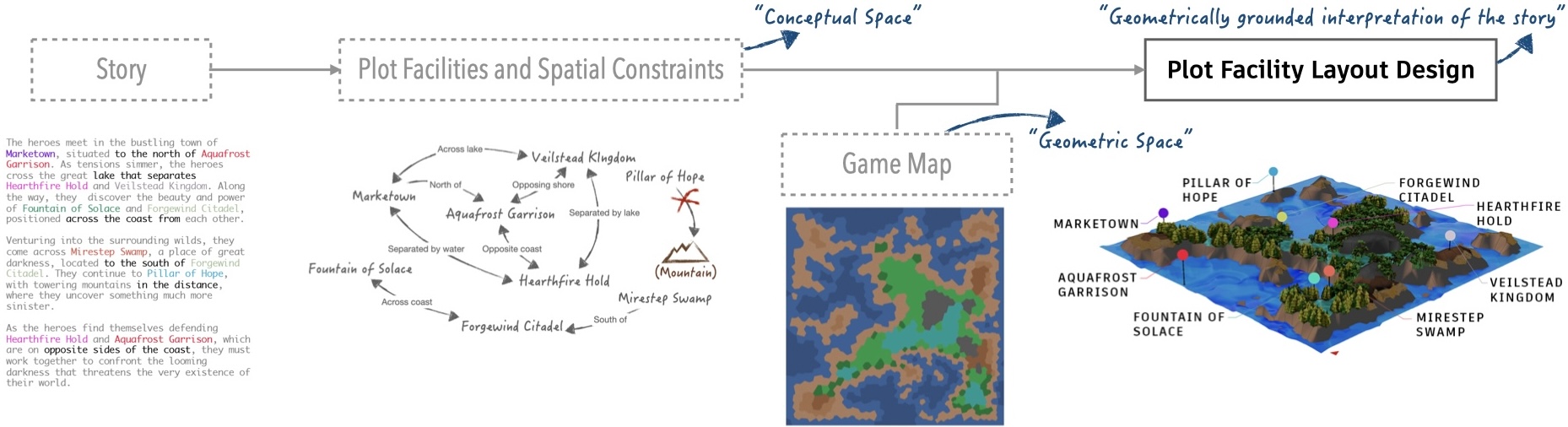}
    \vspace{-1mm}
    \caption{Accommodating a story on a map with a plot facility layout design process}
    \vspace{-3mm}
    \label{fig:pipeline}
\end{figure*}

Figure \ref{fig:pipeline} illustrates the world building approach and plot facility design task that is the focus of the work. We demonstrate our technique with a concrete pipeline using a procedural map generator and story constraints extracted from a textual story description with a large language model.


We consider two different representations of the underlying map: 1) a set of 2D polygons with precise geometric specifications, and 2) a pixel-based 2D image. We present two approaches to automatic plot facility layout design:  an evolutionary computation (EC) based approach through Covariance Matrix Adaptation Evolution Strategy (CMA-ES) for a polygon-based map representation and a Reinforcement Learning (RL) based approach for maps as 2D images. 
We present baseline results for the two approaches. With extensive quantitative and qualitative studies, we show that the EC based approach can rapidly provide accurate solutions for tasks with a practical scale, while the RL based approach has the potential to better support human-in-the-loop map design workflows.


The paper also presents a dataset of facility layout tasks and a Gym-like environment to evaluate different methods and train the RL models. The dataset contains $10,000$ plot facility layout tasks of different scales, involving 12 types of spatial constraints and maximum 60 plot facilities each task, based on a set of procedurally generated terrain maps.\footnote{Our code and dataset are publiclly available at: \url{https://github.com/AutodeskAILab/PlotMap}. } 






In summary, the paper's contribution is threefold: 

\begin{itemize}
    \item We propose plot facility layout design as a novel approach to address the problem of supporting stories with game maps, which is compatible with most story and map generation methods.
    \item We provide a dataset of plot facility layout tasks and a Gym-like environment to experiment with and evaluate different methods for solving the task.
    \item We provide baseline results on a CMA-ES based method, and an RL approach for solving plot facility layout tasks, with discussions on their strengths and weaknesses.
\end{itemize}


\section{Related Work}

\paragraph{Story and Game Map Generation}
Though intertwined, the generation of stories and maps are typically investigated in isolation. A few notable exceptions do tackle them as a single system. Hartsook et al. \cite{hartsook2011toward} proposed a story-driven map generation method where
each event in the plot is associated with a location of a certain environment (e.g., castle, forest, etc.) and a linear plot line is translated to a constraint composed of a sequence of environments. Map generation is formulated as an optimization problem finding a topological structure of the map balancing requirements from a realistic game world and player preferences, subject to plot constraints. Valls-Vargas et al. \cite{valls2013towards} present a procedural method that generates a story and a map facilitating the story at the same time. The problem is formulated as optimization of the topological structure of the map for both playability, and the space of possible stories.

Both Hartsook et al.~\cite{hartsook2011toward} and Valls-Vargas et al.~\cite{valls2013towards} generate grid-based maps consisting of discrete ``scenes'' connected by ``passages''. The map structure is widely used in many classic games such as {\sc Rogue} \cite{rogue} and early {\sc Zelda} series \cite{zelda}. However, many modern RPG games feature seamless world maps with continuous terrains and very few geographical barriers for an immersive open world experience, such as {\sc Elden Ring} \cite{elden_ring} 
and {\sc The Legend of Zelda: Breath of the Wild} \cite{zelda_botw}.  Dormans and Bakkes~\cite{dormans2011generating} use generative grammar based methods for both story (mission) generation and map (space) generation. The story elements are then mapped to spatial elements using heuristics specific to game genre. Our work establishes a mapping between narrative and spatial elements, but with a more general constraint satisfaction process.


\paragraph{Procedural Content Generation}
Procedural Content Generation (PCG) has become an essential component in video games, employed for the algorithmic creation of game elements such as levels, quests, and characters. The primary objectives of PCG are to enhance the replayability of games, reduce the burden on authors, minimize storage requirements, and achieve specific aesthetics \cite{hendrikx2013procedural}, \cite{smelik2009survey}, \cite{kelly2017survey},\cite{Linden2014procedural}. Game developers and researchers alike utilize methods from machine learning, optimization, and constraint-solving to address PCG problems\cite{togelius2011search}. The primary aim of this work is to develop methods for plot facility layout design capable of generalizing across a wide range of environments and constraints. To achieve this goal, we employ a PCG approach to generate a diverse set of maps and constraints as our dataset.
 
\paragraph{Evolutionary Computation for PCG}
Evolutionary computation (EC) reimagines game design challenges as optimization problems solved by evolving solutions to fulfill specific criteria~\cite{dl_pcg}. The flexibility of EC allows for the creative redefinition of design challenges, unbounded by the constraints of precise problem formulations -- for example, by redefining Mario levels as musical compositions and evolved neural networks suggest strategic tile placements~\cite{hoover}.

%

In this work, we employ the Covariance Matrix Adaptation Evolution Strategy (CMA-ES) \cite{cmaes} for solving the plot facility layout task. CMA-ES is an algorithm designed to solve complex, non-linear, and non-convex optimization problems, by evolving a set of solutions, through adjusting a multivariate normal distribution, defined by mean and covariance, based on the success of past solutions. This process involves generating new solutions, evaluating them, and refining the distribution to improve outcomes. CMA-ES dynamically adjusts its search strategy, optimizing the balance between broad exploration and detailed examination of promising areas. It is highlighted for its efficiency, minimal parameter tuning, robustness against local optima and noisy evaluations, and suitability for a wide range of applications. However, its computational and memory demands increase quadratically with problem dimensionality, posing challenges for high dimensional (>1000D) problems.



\paragraph{RL in Modern Video Games}
The popularity of games in AI research is largely attributed to their usefulness in the development and benchmarking of Reinforcement Learning (RL) algorithms~\cite{bellemare2013arcade, jaderberg2019human, vinyals2019grandmaster, berner2019dota}. 
On the other hand, RL can also be used as a design tool for modern games, especially for accessing and testing games. Iskander et al.~\cite{iskander2020reinforcement} developed an RL system to play in an online multiplayer game alongside human players. The historical actions from the RL agent can contribute valuable insights into game balance, such as highlighting infrequently utilized combinations of actions within the game's mechanics. Bergdahl et al.~\cite{bergdahl2020augmenting} used RL as an augmenting automated tool to test game exploits and logical bugs.  Chen et al.~\cite{chen2023emergent} released a multi-agent RL environment to study collective intelligence within the context of real-time strategy game dynamics. To the best of our knowledge, this is the first instance that uses a learning-based approach to accommodate stories on game maps.



\section{Problem Formulation}


\begin{figure*}[t]
    \includegraphics[width=0.9\textwidth]{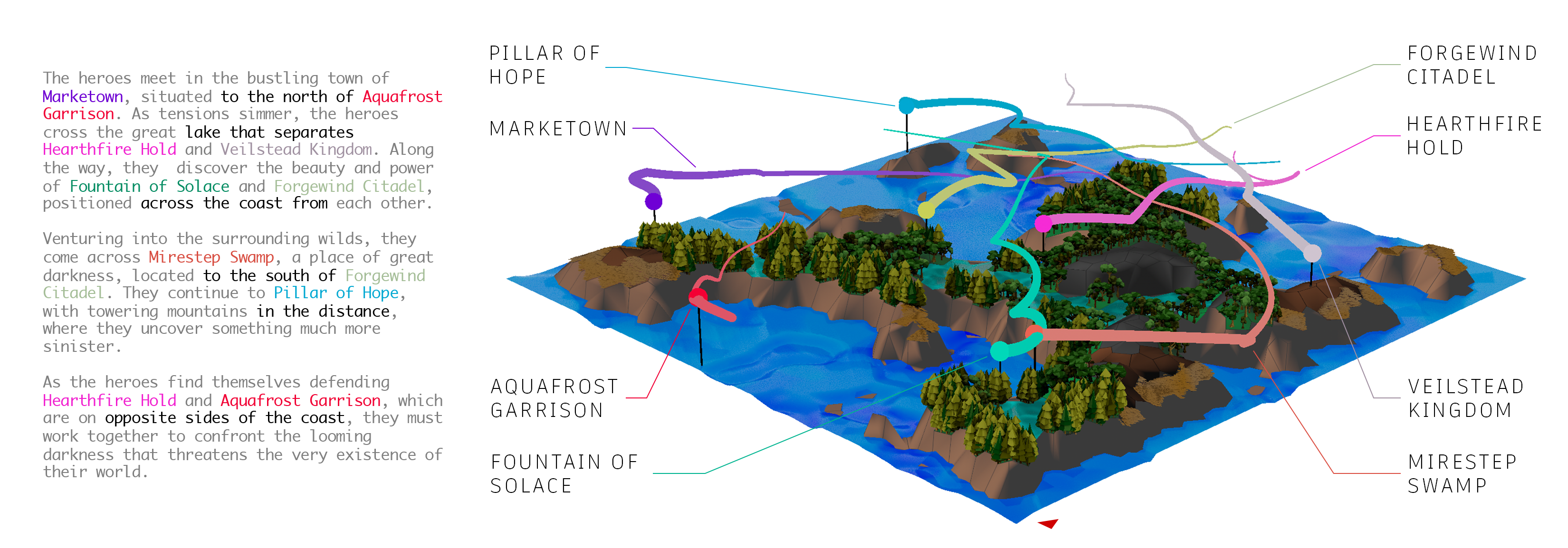}
    \vspace{-4mm}
    \caption{We derive spatial constraints from a story and layout locations mentioned in the story on a map to satisfy the constraints.}
    \vspace{-5mm}
    \label{fig:teaser}
\end{figure*}

Put simply, we would like to assign every location mentioned in a story to an appropriate location on the game map (Fig.~\ref{fig:teaser}). Inspired by Dormans and Bakkes~\cite{dormans2011generating}, we view the story and the geometric layout of a game map as independent of each other, except that the geometric layout should accommodate the story. We introduce the notion of {\em plot facilities}: conceptual ``locations'' mentioned in the story. These ``locations'' are {\em abstract} in the sense that they don't correspond to any concrete geometric locations (yet). For example, the event ``the hero finds an injured dwarf in the forest'' happens at some place. There can be multiple locations on the map where this ``abstract location'' can be ``instantiated'', as long as it does not contradict with the story - in this example it should be inside a forest area.

A set of constraints can be derived from the story for determining whether a particular layout of plot facilities is valid. The set of all plot facilities and the constraints form a conceptual space defined by the story, which is at a higher abstraction level than a concrete game map. The problem is then to assign geometric locations on the map to the plot facilities such that the constraints are satisfied - we call this problem {\em plot facility layout design}. A plot facility layout is essentially a mapping between the conceptual space defined by the story and the geometric space defined by the map.

In the following subsections, we describe our methods for this problem. 
To demonstrate a concrete map generation pipeline, in this study we specifically work with terrain maps, where different regions on the map represent different biomes.



\subsection{Plot Facility Layout Design}

We define a {\em (facility layout) task} as a tuple
\[
\langle \mathcal{F}, \mathcal{T}, \mathcal{C}\rangle
\vspace{-3mm}
\]
where
\begin{itemize}
    \item $\mathcal{F}$ is the set of (plot) facilities. Each facility has an identifier to be referred to by the constraints. 
    \item $\mathcal{T}$ is an underlying terrain map.
    \item $\mathcal{C}$ is a set of spatial constraints over facilities in $\mathcal{F}$ and biome types. 
    \end{itemize}
    An (optimal) solution to a task is an assignment of coordinates to all the facilities in $\mathcal{F}$, so that a maximum number of the constraints in $\mathcal{C}$ are satisfied considering their relations with each other and the biomes in $\mathcal{T}$.

In the subsequent sections, we consider two different representations of the map $\mathcal{T}$: 1) as a set of 2D polygons on the map, each associated with a biome type (e.g., {\tt OCEAN}, {\tt PLAINS}, etc.), or 2) as a 2D image, with color-coded regions representing biomes. We also consider two different representations of the constraints in $\mathcal{C}$: 1) as relational expressions such as $\texttt{AcrossBiomeFrom}$ $(\texttt{OCEAN}, \texttt{fordlen\_bay}, \texttt{snapfoot\_forest})$, or 2) as natural language (NL) utterances, such as ``Fordlen Bay and Snapfoot Forest are separated by the ocean.''. 
\section{Task Dataset Generation}\label{sec:dataset_generation}

We generate a dataset of 10,000 facility layout tasks for training and evaluation. Each task requires arranging the layout of minimum $10$ to maximum $60$ plot facilities on top of a procedurally generated map, w.r.t. a set of maximum $90$ spatial constraints. Maps consist of $9$ biome types and the constraints are generated based on $12$ constraint types. 


\subsection{Map Generation}\label{sec:map_generation}

We employ a procedural map generation approach adapted from Patel \cite{patel2010polygonal} and its implementation by Dieu et al. \cite{Polygonal-Map-Generation-for-Games}.
Initially, a grid of Voronoi polygons is created from random points, followed by Lloyd relaxation to ensure even distribution. Coastline generation uses a flooding simulation to designate ocean, coast, lake, and land tiles based on water edge proximity and neighboring tiles. Elevation is determined by distance from the coast, with a distribution that emphasizes smoother terrain with fewer high points. Rivers flow from mountains to the nearest water body, while moisture levels are influenced by proximity to freshwater sources. Finally, the biome of each polygon is assigned based on a combination of moisture and elevation.

To generate a dataset, 100 maps with 1000 polygons each are produced.
These maps are also converted to RGB images, suitable for input to neural network-based RL agents.


\definecolor{OCEAN}{RGB}{0, 191, 255}
\definecolor{LAKE}{RGB}{65, 105, 225}
\definecolor{COAST}{RGB}{245, 245, 220}
\definecolor{MOUNTAIN}{RGB}{112, 128, 144}
\definecolor{FOREST}{RGB}{34, 139, 34}
\definecolor{HILLS}{RGB}{189, 183, 107}
\definecolor{WOODED_HILLS}{RGB}{0, 100, 0}
\definecolor{PLAINS}{RGB}{245, 222, 179}
\definecolor{DEEPOCEAN}{RGB}{30, 144, 255}


\subsection{Constraint Generation}\label{sec:constraint_generation}

\begin{figure}
    \includegraphics[width=0.45\textwidth]{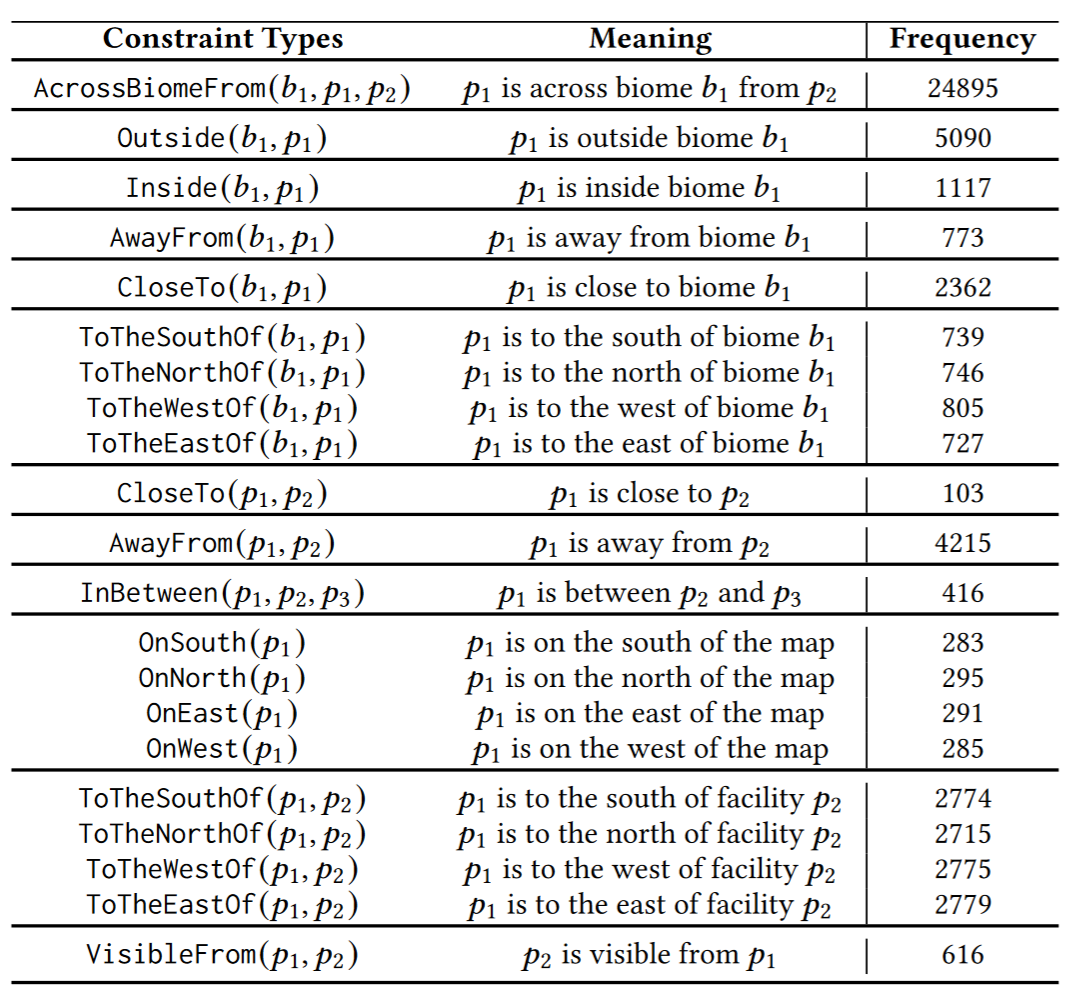}
    \vspace{-1mm}
    \caption{Constraint types included in the $10,000$-task dataset (each $p_i$ represents a plot facility).  A constraint type $ConstraintType(b_1, \dots, b_m, p_1, \dots, p_n)$ is {\em instantiated} to become a constraint by substituting each of $b_1, \dots, b_m$ with a biome type, and each of $p_1, \dots, p_m$ with a plot facility id ($m \geq 0, n\geq 0$).}
    \vspace{-3mm}
    \label{fig:constraint_type_table}
\end{figure}

Facility layout tasks are generated by associating a set of random constraints to randomly chosen map. Figure~\ref{fig:constraint_type_table} lists the constraint types and their frequency in the dataset. The constraint types were selected to cover common geometric relations between points (facilities) and polygons (biomes), and they are represented as unary, binary and ternary relations.  

For each constraint type, we define a heuristic function for evaluating an existing facility layout w.r.t. any instantiation of the constraint type. The function returns a real number in $[0.0, 1.0]$ with $1.0$ meaning fully satisfied and $0.0$ completely unsatisfied.\footnote{E.g., closeTo(x,y) is negatively correlated to the distance between x and y, and reaches 1 when their distance is less than a certain threshold.} These functions are used to check if the randomly generated constraints are satisfied by a random layout and to compute the objective value or reward. We associate with each constraint a natural language utterance (e.g., ``Fordlen Bay and Snapfoot Forest are separated by the ocean.'') for training the RL model. Tasks are then generated following Algorithm \ref{alg:task_generation}. Note that, as the constraints are extracted from an example layout, they are guaranteed to be solvable.

\begin{algorithm}
\caption{Facility Layout Task Generation}
\label{alg:task_generation}
{\footnotesize
\SetAlgoLined
\KwIn{A set of map $MAP$, maximum number of facilities $N$, a set of constraint types $\mathcal{CT}$, minimum and maximum number of constraints $M_1$ and $M_2$}
\KwOut{A facility layout task $\langle F, T, C\rangle$}
1. Randomly sample a map $T$ from $MAP$\;
2. Randomly assign a location to facilities $obj_1\dots obj_N$ on the map $T$\;
3. For each constraint type in $\mathcal{CT}$, generate all possible instantiations of it w.r.t. $obj_1\dots obj_N$ and biome types. Evaluate each of them against the current map, adding the true ones to a set $C^\prime$ \;
4. Sample a set of statements from $C^\prime$ sized between $M_1$ and $M_2$, obtaining $C$\;
5. For each statement in $C$, use large language model such as GPT~\cite{brown2020language} to rephrase it with a natural language sentence, resulting in a set of NL utterances $C^{NL}$\;
6. \Return task $\langle \{obj_1\dots obj_N\}, T, C^{NL}  \rangle$.
}
\end{algorithm}

\begin{figure}
    \centering 
    \includegraphics[width=0.5\textwidth]{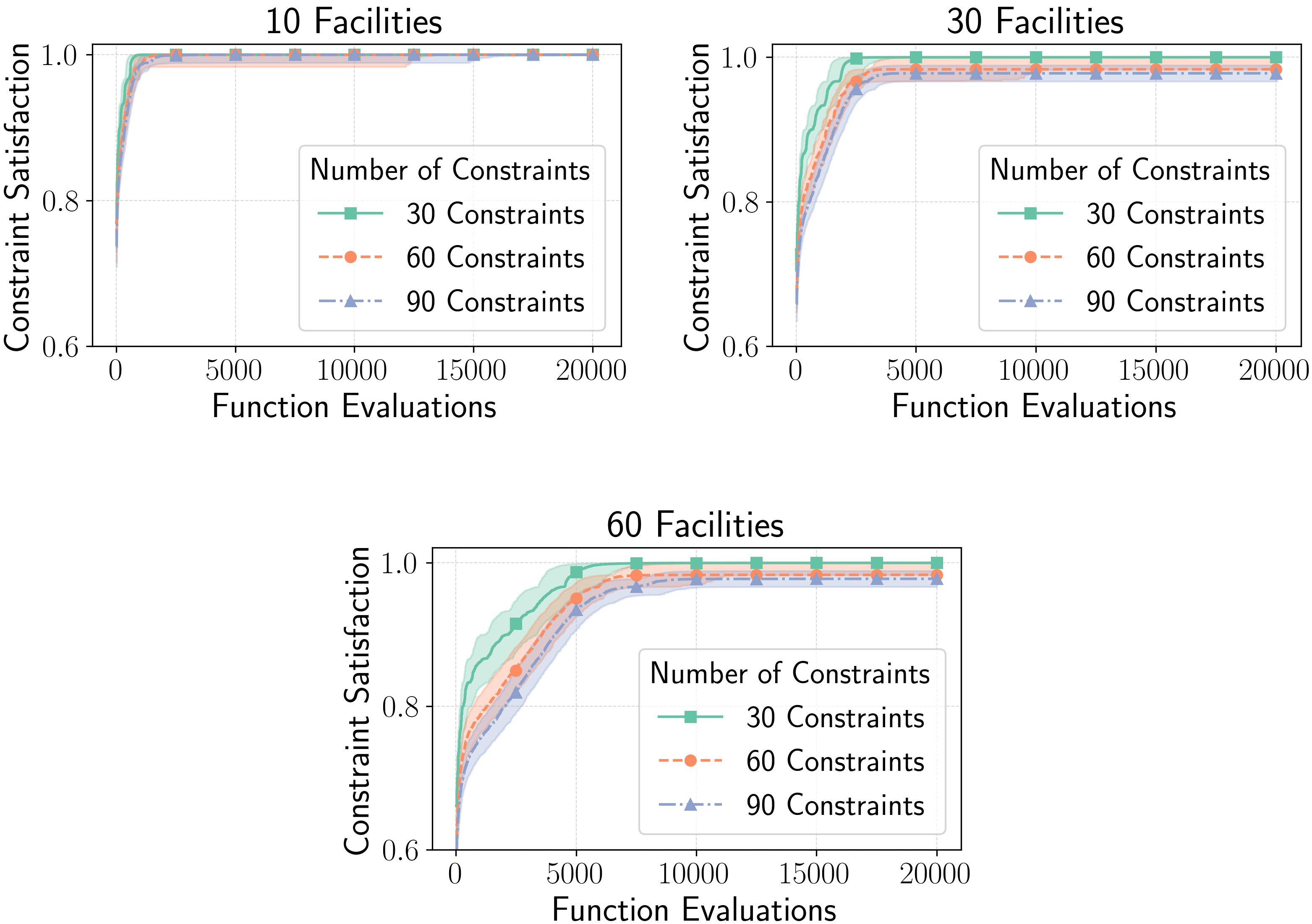}
    \vspace{-6mm}
    \caption{
    Layout evaluations to reach a level of constraint satisfaction with 10, 30 and 60 facilities, and 30, 60, and 90 constraints to satisfy. Median satisfaction over 1000 different tuples of terrain and constraints, shaded regions indicate 1st and 3rd quartiles. Adaptive population sizes and cause generation times to vary from 0.25 to 0.75 seconds. However, performing a run with a budget of 20k evaluations consistently takes 2-5 minutes.
    }
    \vspace{-3mm}
    \label{fig:cmaes}
\end{figure}

\vspace{-5mm}
\section{Evolutionary Computation (CMA-ES) Approach}
\label{sec:cma-es}
When the map is represented as polygons and the constraints as relational expressions, the facility layout problem can be approached as a black-box optimization problem where the goal is to find facility positions, represented by (x, y) coordinates, to maximize constraint satisfaction. Each coordinate of each facility is a single dimension of search (e.g., the location of 10 facilities is a 20 dimensional problem). The search is guided only by feedback from calls to an objective function that tests the satisfaction of the entire layout. We use a common variant of CMA-ES, the Increasing Population Size CMA-ES~\cite{auger2005restart} which restarts with a population size that doubles after each restart, based on predefined stagnation criteria. 

\subsection{Quantitative Evaluation}
We evaluate the performance of CMA-ES on task configurations with 10, 30, and 60 facilities, each under conditions of 30, 60, and 90 constraints. For each configuration, we tested 1000 examples generated, as described in Section \ref{sec:dataset_generation}, allowing us to explore the impact of an increasing number of constraints on the optimization efficiency of a fixed number of facilities.

The plots in Figure \ref{fig:cmaes} show that CMA-ES performs extremely well in low-dimensional settings, rapidly finding nearly perfect solutions even as the number of constraints grows. However, as the number of facilities grows, the number of evaluations required to reach a solution increases. The performance drop due to more constraints is low, highlighting the benefits of the black-box approach. 


\section{A Reinforcement Learning Approach}
\label{sec:rl}

Our CMA-ES based approach relies on repeated computation of the constraint satisfaction degree at solving time, which is enabled by the polygon-based representation of the maps.
For many use cases, such a polygon-based representation of the map may not be available, for example when the map is hand-drawn or a photo of a physical spatial structure. Furthermore, the constraints may not be characterizable by closed-form arithmetic expressions. They may be from aesthetic considerations or subjective preferences that can only be statistically captured. For example, a game designer may want a ``balanced'' map layout. This motivates us to explore a learning-based approach for plot facility layout design.

In this section, we present a Reinforcement Learning (RL) method as a preliminary exploration. We employ a decision-making agent to optimize facility layout on a 2D pixel map image, considering the constraints of the story. This approach eliminates the need for solving-time constraint satisfaction computation, and allows for constraints to be expressed in natural language, enhancing the system's flexibility and applicability to a wider range of design scenarios.

\subsection{RL Formulation}

\label{sec:rl_formulation}
Each plot facility layout task can be viewed as sequentially moving a set of plot facilities $\mathcal{F}$ on a map, thus we define the plot facility layout design as a sequential Markov decision process (MDP), with the following elements:

\begin{itemize}
  \item Each state $s \in S$ consists of three modalities: 1) a pixel-based image representing the map, 2) a real-valued vector representing essential information of the plot facilities, and 3) a group of 
  constraints. 
  \item Each action $a \in A$ is a 2d vector of real-valued $[\Delta x, \Delta y]$ for one plot facility. In each round, plot facilities are moved one at a time in a fixed order, with the state and rewards updated after each movement. We set the range of $\Delta x$ and $\Delta y$ to be small steps so that we are simulating a concurrent movement of the facilities at a macro-level. 
  \item The reward (for each step) $r_t$ is $+1$ when all constraints are satisfied, and is the average satisfaction score from all constraints minus $1$ when partial constraints are satisfied:

\[
r_t = \left\{
  \begin{array}{ll}
    1, & \text{if all constraints are satisfied} \\
    \frac{1}{n} \sum_{i=1}^{n} s_i - 1, & \text{otherwise}
  \end{array}
\right.
\]
  
  where $n$ is the number of constraints and $s_i$ is the satisfaction score for each constraint. The satisfaction score for each type of constraint is within $[0, 1]$ and defined based on the heuristic functions described in Section \ref{sec:constraint_generation}. The range of the reward $r_t$ is $[-1, 0] \cup \{1\}$.
  \item The transition function is deterministic, where $s_{t+1} = f(s_t, a_t)$. 
  \item Each episode is terminated when all the constraints are satisfied or at 200 timesteps.
\end{itemize}

We train an RL agent to learn an optimal policy $\pi_\theta$ in order to maximize the expected discounted rewards: 

\begin{equation}
\label{eq1}
\max_{\pi_\theta}\mathbb{E}_{\tau\sim\pi_\theta}\left[ \sum_{t=0}^{T} \gamma^{t} r(s_t, a_t) \right],
\end{equation}

where trajectory $\tau = (s_0, a_0, s_1, a_1, ..., s_T, a_T)$, $\theta$ is the parameterization of policy $\pi$, and $\gamma$ is the discounted factor. Our goal is to train a general RL agent to be able to find solutions to any arbitrary task.

\section{Experiments \& Analysis}

\subsection{Experiment Setup}
\label{sub:exp_details}

Our state space includes $3$ types of inputs:

\begin{itemize}
    \item a map encoded as a ($42\times 42\times 3$) RBG image.
    \item constraints represented by natural language utterances or relational expressions, depending on the embedding strategies described in the following paragraph.
    \item Each plot facility's information is represented by a vector, consisting of its position $[x, y]$ on the map, a binary motion indicator signifying if it is its turn to move or not, and a unique identifier.   
\end{itemize}

In all of our experiments, we have set a limit of 10 for both the number of plot facilities and the number of constraints. All of our policies are trained using Proximal Policy Optimization (PPO) \cite{schulman2017proximal}. To handle the multi-modal observation space, we employ pre-trained models to independently extract embeddings from the map (image) and constraint (natural language) inputs. These embeddings are subsequently concatenated with the informational vector and provided as inputs to the policy network. Specifically, we design two strategies for deriving the embeddings:

\begin{itemize}
    \item \textit{NL-based}: using ResNet~\cite{he2016deep} for maps and SentenceTransformer~\cite{reimers2019sentence} for the NL representation of constraints. The state dimensions are $4,392$, consisting of 512 dimensions for map, $40$ for plot facilities, and $384 \times 10$ for constraints. 
    \item \textit{Relation-based}: using ResNet for maps, and each constraint, as a relational expression, is encoded as a one-hot vector, representing the constraint type, followed by three one-hot vectors indicating the specific plot facilities to instantiate the constraint with. The state dimensions are $1,782$, consisting of $512$ dimensions for map, $40$ for plot facilities, and $123\times 10$ for constraints.
\end{itemize} 

\subsection{Quantitative Evaluation} 

\begin{table*}[tbh]%
\small
\caption{Success rate (\%) comparison between a random agent and the two proposed baseline methods across task sizes and conditions. Each baseline is assessed under two conditions: 1) random initialization on the same training tasks, and 2) random initialization on 100 unseen tasks. For the 100-task datasets, we examine three combinations of maps and constraints.}
\vspace{-3mm}
\label{tab:varying-tasks}
\begin{center}

\begin{center}

\begin{tabular}{c|c|cccccc}

  \toprule
  \multirow{4}{*}{\textbf{Method}} & \multirow{4}{*}{\textbf{Evaluation Condition}} & \multicolumn{6}{c}{\textbf{Success Rate (\%)}}\\
   & & 1 task & 5 tasks & 50 tasks & \multicolumn{3}{c}{100 tasks} \\
   & & & & & 10 maps +  & 1 map + & 100 maps + \\
  & & & & &  100 sets of constrs & 100 sets of constrs & 1 set of constrs \\

  \midrule
  Random Agent & & 23.9 & 30.3 & 30.5 & 30.9 & 34.5 & 38.0 \\
  
  \midrule
  \multirow{2}{*}{NL-based} & rand init & 84.1 & 48.2 & 38.5 & 38.4 & 42.8 & 79.2 \\ 
  & rand init + unseen tasks & 34.8 & 33.3 & 35.1 & 38.0 & 39.9 & 70.4\\
  
  \midrule
  \multirow{2}{*}{Relation-based} & rand init & 100.0 & 55.1 & 46.1 & 38.5 & 48.1 & 77.5 \\
  & rand init + unseen tasks & 32.2 & 29.6 & 32.7 & 36.1 & 40.1 & 68.9\\
  
  \bottomrule

\end{tabular}

\end{center}

\vspace{-5mm}
\end{center}
\end{table*}%

We examine the performance of our two proposed embedding strategies across four small task sets. For the datasets comprising 100 tasks, we explore the influence of maps and constraints on generalization. These results are presented in Table~\ref{tab:varying-tasks}.

For each dataset, we evaluate our trained policies under two conditions: \textit{rand init}, which refers to evaluating the policies with the same task sets used for training but under different initial positions; and \textit{rand init + unseen tasks}, which reports the success rates when the policies, each trained on their respective task set, are tested on 100 unseen tasks (with different maps and constraints). Success is considered only when all constraints are satisfied. We calculate the success rates over 1,000 rollouts, with each rollout taking approximately 5 seconds to complete. 

Table~\ref{tab:varying-tasks} shows that both baselines perform exceptionally well on tackling a single hard task. The success rate for both baselines decreases as the size of the task set increases, exhibiting a deficiency in generalization capability.  


To investigate further the factors impeding generalization, we study the influence of maps and constraints by training on three distinct sets of 100 tasks as depicted in the last three columns in Table~\ref{tab:varying-tasks}. The substantial disparity between varying sets of constraints and one fixed set of constraints indicates that constraints pose a greater challenge to generalization, regardless of the encoding method employed. 


\section{Qualitative Results}\label{sec:qualitative_result}

\begin{figure}
    \centering
    \includegraphics[width=0.35\textwidth]{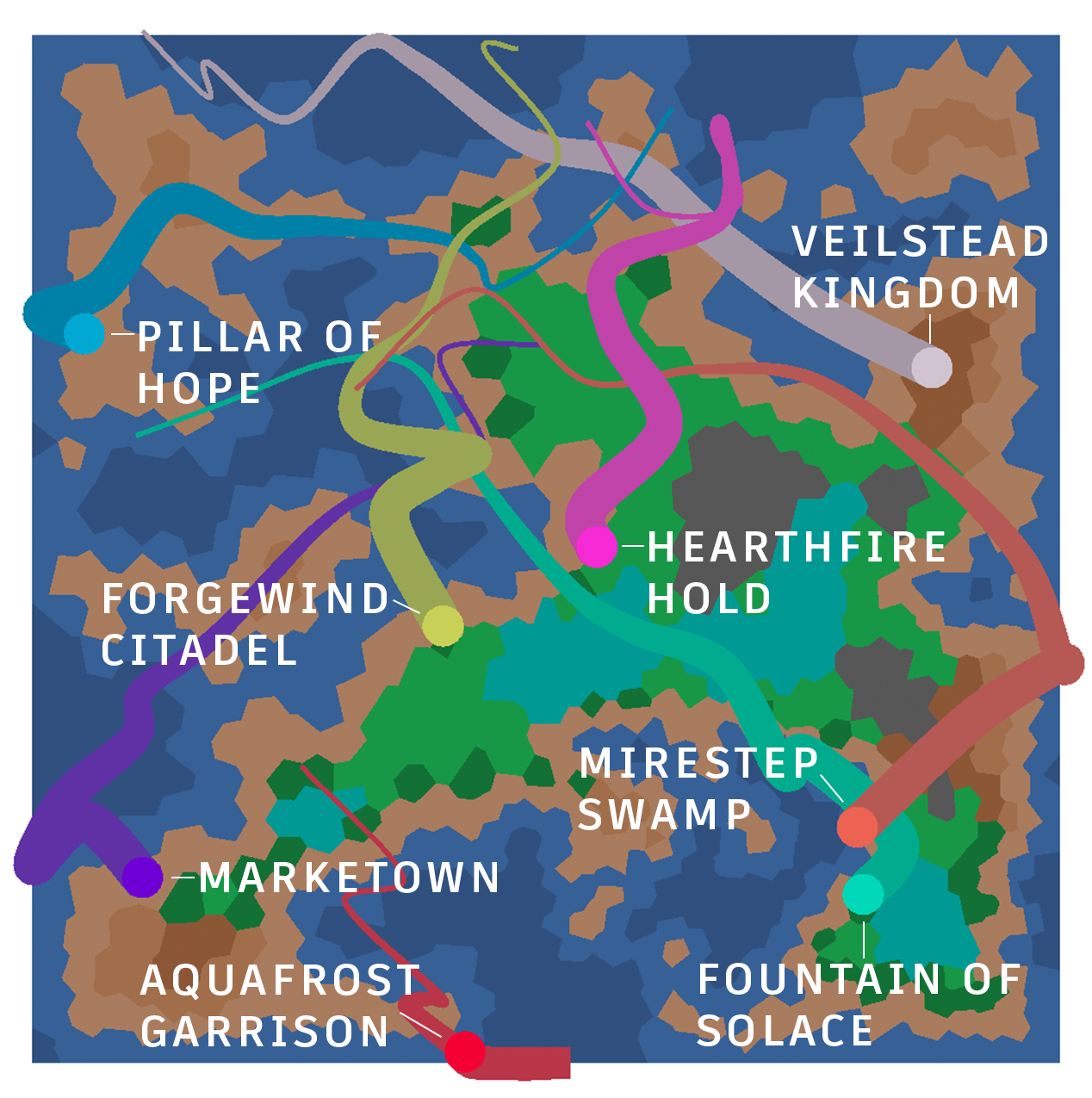}
    \vspace{-3mm}
    \caption{\textbf{Cooperative behavior to satisfy constraints}: \texttt{Marketown} and \texttt{Veilstead Kingdom} must be across a lake from each other, while \texttt{Aquafrost Garrison} must be to the south of \texttt{Marketown}. \texttt{Aquafrost Garrison} was initially to the south of \texttt{Marketown} but as \texttt{Marketown} moves south to be across the lake from \texttt{Veilstead Kingdom}, \texttt{Aquafrost Garrison} moves even further south to continue satisfying its south of \texttt{Marketown} constraint.}
    \vspace{-3mm}
    \label{fig:compound_behavior}
\end{figure}

\begin{figure*}
    \includegraphics[width=0.95\textwidth]{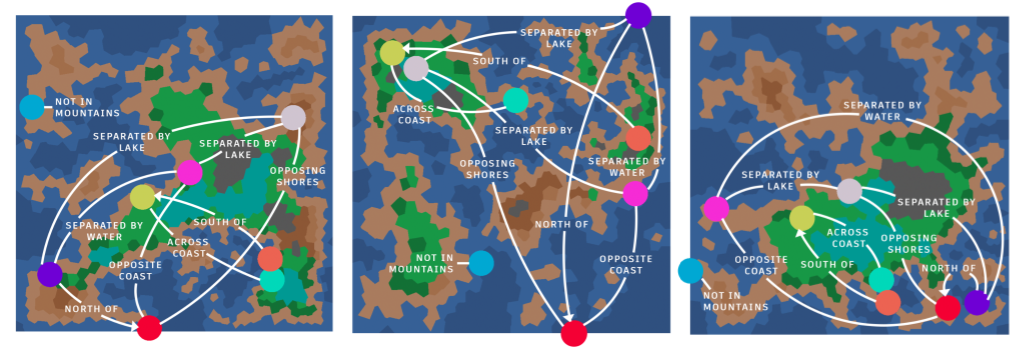}
    \vspace{-5mm}
    \caption{Same story accommodated on 3 different maps}
    \vspace{-2mm}
    \label{fig:task5_alternative_maps}
\end{figure*}

In this section, we demonstrate a complete pipeline of the system, and highlight insightful behaviors of the CMA-ES and RL method through specific examples.

Figure \ref{fig:teaser} illustrates a complete story-to-facility-layout pipeline. From the story on the left, $8$ plot facilities and $6$ constraints are extracted with a pre-trained large language model. 


The map on the right shows the resulting plot facility layout, along with motion trails indicating traces to the final locations from rolling out a trained RL model. The RL models demonstrate interesting cooperative behaviors (Fig. \ref{fig:compound_behavior}).


   
  




In Figure \ref{fig:task5_alternative_maps}, we demonstrate the same story accommodated on 4 different maps. We observe that the same set of plot facilities can still maintain their relative spatial relations on completely different geometric layouts, which aligns with the perspective described in \cite{dormans2011generating}: ``The same mission (story) can be mapped to many different spaces, and one space can support multiple different missions (stories)''. This capability enables designers to envision various interpretations of unspecified story details and potential story progressions.

\begin{figure}
    \centering
    \includegraphics[width=0.50\textwidth]{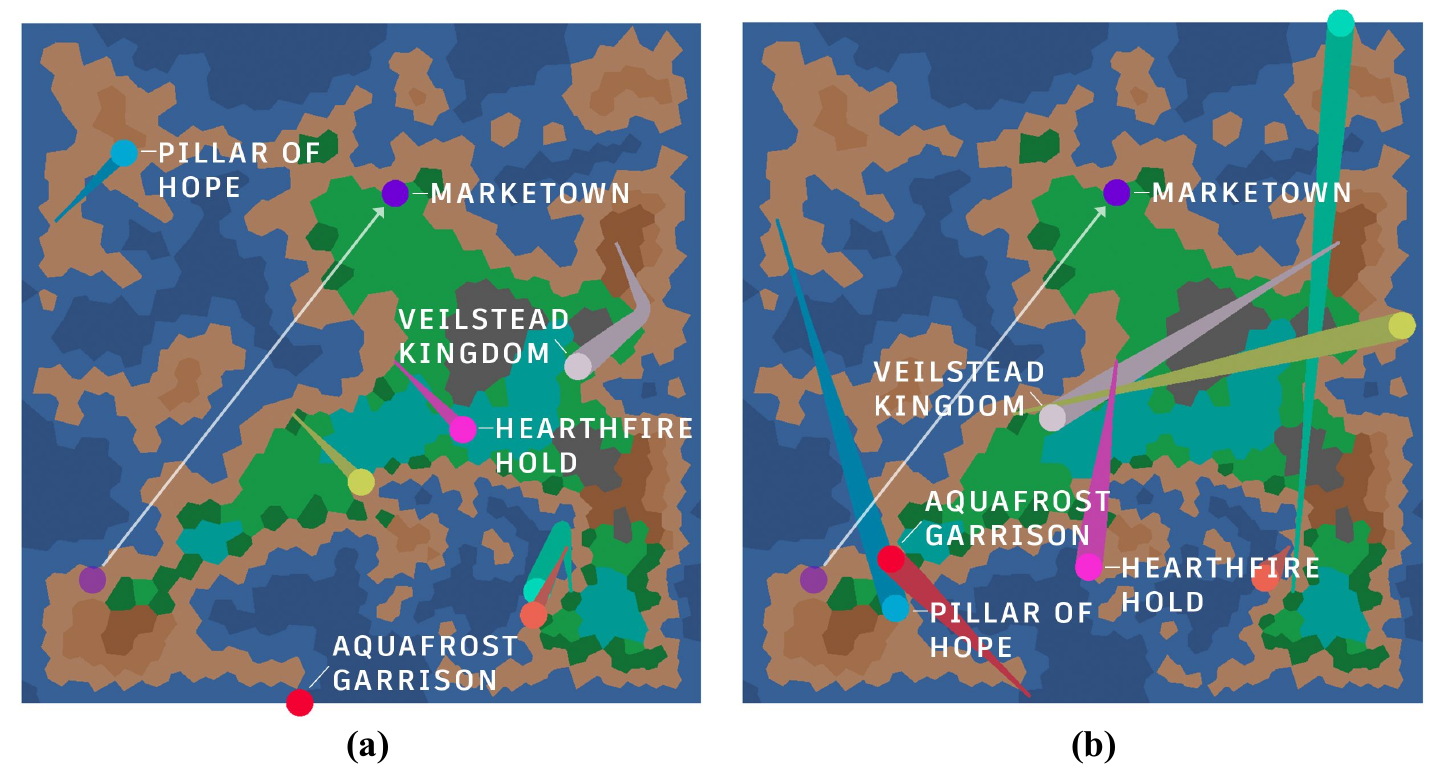}
    \vspace{-8mm}
    \caption{Plot facility re-adaptation after the user moves \texttt{Marketown} to a different location (a) with RL, and (b) with CMA-ES. (a) can be seen as an adjustment of the previous layout while (b) changes into a completely different layout.}
    \label{fig:user_intervention}
\end{figure}

The RL method provides smooth traces from the starting positions to the final layout, enabling an interactive interface for adjusting existing layout solutions. In Figure \ref{fig:user_intervention} (a), we show that our RL policies support real-time re-adaptation after human intervention. Specifically, after we manually change the location of \texttt{Marketown} to the northeast part of the map, \texttt{Veilstead Kingdom} and \texttt{Hearthfire Hold} can adjust their locations to continue to be across the lake from \texttt{Marketown}, while \texttt{Aquafrost Garrison} stays at the same location, so all of the constraints are still satisfied. In contrast, Figure \ref{fig:user_intervention} (b) shows the the result obtained by CMA-ES, again with a manual change of the location of \texttt{Marketown}. For the rest of the facilities, we start with their current location as the initial guess, and re-run the CMA-ES process until reaching $1.0$ satisfaction. As can be seen, the configuration of the rest of the facilities has been drastically changed, with no obvious connection with the previous layout.

\section{Limitation and Future Work}

The scalability of CMA-ES is limited by the dimensionality of the problem and the complexity of the objective function.
The quadratic scaling of the size of the covariance matrix dramatically reduces the speed of CMA-ES with a large number of parameters, and CMA-ES is not often used in cases where it exceeds 1000. This translates to 500 locations as each location is represented by two parameters (x,y). The 2011 release {\sc Skyrim} had 670\footnote{\url{https://screenrant.com/skyrim-how-many-locations-big-map-size/}}, the more recent {\sc Elden Ring} had a map twice the size with locations at higher density\footnote{\url{https://screenrant.com/skyrim-map-elden-ring-lands-between-big-size}}. In the truly large scale cases where CMA-ES would be most useful it is also the least practical.

Our dataset generation procedure resulted in an imbalanced dataset, as shown in Figure \ref{fig:constraint_type_table}, which has led to unexpected behaviors and failure cases of the trained RL models. 
For example, in the dataset there is a significantly higher proportion of $\texttt{AcrossBiomeFrom}$ constraint. This had led to a strong tendency of the RL model to move facilities to the edge of the map, since two plot facilities on different edges of the map are likely to be across several different biomes from each other, despite human designers may prefer the facilities to be close to the biome.\footnote{For example, ``A and B are across a lake'' generally implies that A and B are on the shore of the lake.} The RL model also tends to fail to satisfy constraint types with fewer occurrences in the dataset (such as $\texttt{CloseTo}(p_1, p_2)$). The behavior of the CMA-ES method is not affected by the imbalance of the dataset. However, it could still produce undesirable maps due to the inconsistency between the human designer's interpretation and the computational implementation of a constraint.

The use of hand-crafted objective functions present challenges in both accurately reflecting the preferences of human designers and providing the right signal for training. A promising solution for better understanding the kinds of solutions which designers prefer is to adopt a statistically trained preference model such as used in Reinforcement Learning from Human Preference, which offers potentially more accurate reflections of human intent~\cite{christiano2017deep}. Moreover, we considered the satisfaction of all constraints as the benchmark for successful task resolution. In practice, a suboptimal solution that satisfies most of the constraints might be acceptable, and situations where the constraints are unsatisfiable are completely possible. In these cases knowledge of designer preferences, or a mixed-initiative approach \cite{Yannakakis2014MixedinitiativeC, smith2010tanagra, cook2021danesh} which allowed editing of the map, would allow desirable solutions to be found.


The formulation, scalability, and embedding strategies of the RL approach all presented challenges. Employing a single RL agent to manage all global information and plot facilities is a bottleneck, hindering performance in larger settings. A distributed RL model, where each facility operates as an independent agent could allow the system to better scale. Achieving a high level of generality within the RL model, akin to adapting to various game levels or rules, poses a significant challenge. 
Drawing parallels with RL applications in gaming, where agents must learn to generalize to varied conditions, a curriculum learning approach~\cite{portelas2021automatic} could offer a solution by dynamically generating maps and constraints while progressively increasing scenario complexity and diversity, thereby fostering the model's generalization capabilities.

\section{Conclusion}

In this work we introduced new methods to support stories with game maps through an automated plot facility layout design process. We demonstrated that this approach allows us to utilize existing story and map generation techniques, and visualize the spatial implications of a story and the narrative potential of a map. We present two methods for solving plot facility layout tasks: an EC based approach through CMA-ES, and an RL based approach, suitable for different representations of the underlying maps. With extensive quantitative and qualitative studies, we show that the EC based approach can rapidly provide accurate solutions for tasks with a practical scale, while the RL based approach has the potential to better support human-in-the-loop map design workflows. The concept of plot facility layout design has potential in many game design applications, such as map design, playtime quest generation/adaptation and story debugging; but also potential applications in other domains involving spatial layouts subject to constraints, such as the design of large office buildings or manufacturing plants. 

\bibliographystyle{IEEEtran}
\bibliography{bibliography}


\end{document}